\documentclass[letterpaper, 10 pt, conference]{ieeeconf}
\IEEEoverridecommandlockouts    
\usepackage{graphics}           
\usepackage{times}              
\usepackage{amsmath}            
\usepackage{amssymb}            
\usepackage{graphicx}
\usepackage{algorithm}
\usepackage[noend]{algpseudocode}
\usepackage{booktabs}
\usepackage{color}
\definecolor{instructioncolor}{rgb}{.5,.5,.5}

\usepackage[font=small]{caption}

\def\eqref#1{Eq.~(\ref{#1})}

\makeatletter
\usepackage{xspace}
\DeclareRobustCommand\onedot{\futurelet\@let@token\@onedot}
\def\@onedot{\ifx\@let@token.\else.\null\fi\xspace}

\makeatother

\usepackage{array}
\newcolumntype{L}[1]{>{\raggedright\let\newline\\\arraybackslash\hspace{0pt}}m{#1}}
\newcolumntype{C}[1]{>{\centering\let\newline\\\arraybackslash\hspace{0pt}}m{#1}}
\newcolumntype{R}[1]{>{\raggedleft\let\newline\\\arraybackslash\hspace{0pt}}m{#1}}

\usepackage{hyperref}   
\usepackage{xcolor}     
\usepackage{multirow}
\usepackage{graphicx}
\usepackage{cleveref}   
\crefname{section}{Sec.}{Secs.}
\Crefname{section}{Section}{Sections}
\Crefname{table}{Table}{Tables}
\crefname{table}{Tab.}{Tabs.}
\crefformat{figure}{Fig.~#2\textcolor{red}{#1}#3}
\crefformat{table}{Tab.~#2\textcolor{red}{#1}#3}
\newcommand\scalemath[2]{\scalebox{#1}{\mbox{\ensuremath{\displaystyle #2}}}}
\hypersetup{
    colorlinks=true,
    linkcolor=blue,
    citecolor=blue,
    filecolor=blue,
    urlcolor=blue,
}

\title{\LARGE \bf Efficient Multimodal 3D Object Detector via Instance-Level \\ Contrastive Distillation}

\author{Zhuoqun Su \and Huimin Lu \and Shuaifeng Jiao \and Junhao Xiao \and Yaonan Wang \and Xieyuanli Chen
\thanks{$^1$Z. Su, H. Lu, S. Jiao, and X. Chen are with the College of Intelligence Science and Technology, and the National Key Laboratory of Equipment State Sensing and Smart Support, National University of Defense Technology. $^2$Y. Wang is with the
College of Electrical and Information Engineering, Hunan University, China.}
  \thanks{Corresponding author: Xieyuanli Chen (xieyuanli.chen@nudt.edu.cn).}
  \thanks{This work was supported by the National Science Foundation of China under Grant 62403478 and 62203460, Young Elite Scientists Sponsorship Program by CAST (No. 2023QNRC001), as well as Major Project of Natural Science Foundation of Hunan Province under Grant 2021JC0004.
  }
}

\begin{document}
\maketitle
\thispagestyle{empty}
\pagestyle{empty}

\begin{abstract}
Multimodal 3D object detectors leverage the strengths of both geometry-aware LiDAR point clouds and semantically rich RGB images to enhance detection performance. However, the inherent heterogeneity between these modalities, including unbalanced convergence and modal misalignment, poses significant challenges. Meanwhile, the large size of the detection-oriented feature also constrains existing fusion strategies to capture long-range dependencies for the 3D detection tasks. In this work, we introduce a fast yet effective multimodal 3D object detector, incorporating our proposed Instance-level Contrastive Distillation (ICD) framework and Cross Linear Attention Fusion Module (CLFM). ICD aligns instance-level image features with LiDAR representations through object-aware contrastive distillation, ensuring fine-grained cross-modal consistency. Meanwhile, CLFM presents an efficient and scalable fusion strategy that enhances cross-modal global interactions within sizable multimodal BEV features. Extensive experiments on the KITTI and nuScenes 3D object detection benchmarks demonstrate the effectiveness of our methods. Notably, our 3D object detector outperforms state-of-the-art (SOTA) methods while achieving superior efficiency. The implementation of our method has been released as open-source at: \href{https://github.com/nubot-nudt/ICD-Fusion}{https://github.com/nubot-nudt/ICD-Fusion}.
\end{abstract}

\section{Introduction}
\label{sec:intro}
Autonomous driving research has extensively explored the integration of cameras and LiDAR sensors, leveraging their complementary strengths to enhance 3D object detection performance. Camera images capture rich semantic details, such as color and texture, whereas LiDAR point clouds provide precise depth and geometric information. However, the inherent heterogeneity between these modalities, stemming from differences in sparsity and perspective, poses significant challenges for multimodal 3D object detection.

\begin{figure}[t]
  \centering
  \includegraphics[width=0.95\linewidth]{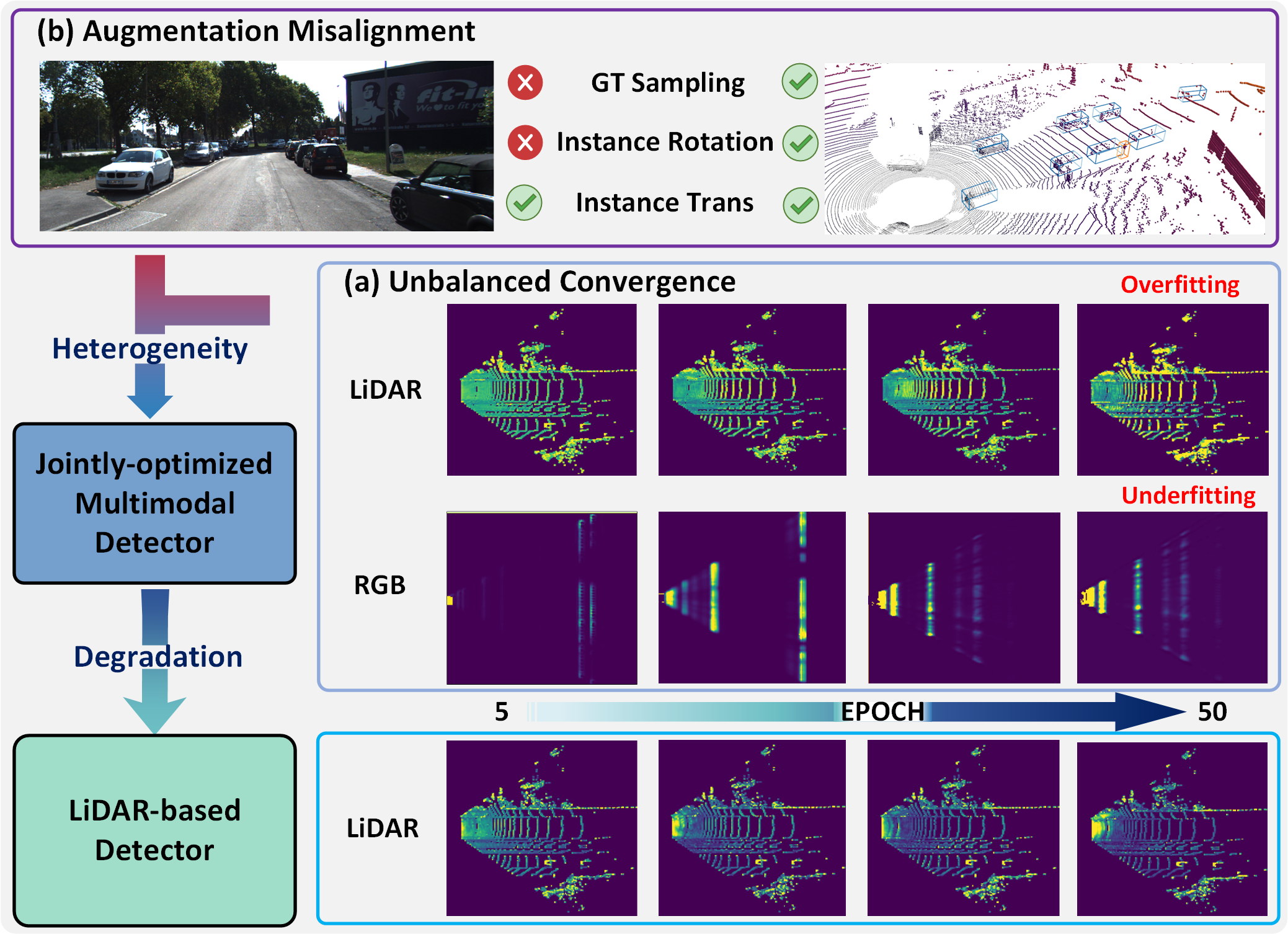}
  \caption{(a) During the early stages of training, the dual-branch encoder tends to over-rely on LiDAR feature due to its faster convergence, leading to severe trailing effects in the RGB-derived BEV feature. (b) LiDAR-based augmentation techniques cannot easily applied to a dual-branch encoder, as ensuring spatial and semantic consistency in RGB images remains challenging when attempting equivalent transformations.}
  \label{fig:motivation}
  \vspace{-0.3cm}
\end{figure}

Multimodal learning generally outperforms its unimodal counterpart in perception tasks. For example, BEVFusion~\cite{liu2023icra} utilizes two separate encoders to independently project LiDAR and image data into a shared BEV space. This decoupled dual-branch approach preserves modality-specific information and subsequently fuses the BEV representations via a simple ConvFusion~\cite{waeijen2021acce} layer, resulting in notable performance improvements.
However, a counterintuitive observation when applying such a framework to 3D object detection is that dual-branch multimodal detectors sometimes underperform relative to LiDAR-only detectors. 

To investigate this, we conducted comparative experiments using a LiDAR-based 3D detector~\cite{deng2021aaai} and a dual-branch multimodal detector~\cite{liu2023icra}.
As shown in~\cref{fig:motivation}(a), BEV features from different modalities exhibit varying degrees of convergence in a jointly optimized multimodal 3D detector due to inherent cross-modal heterogeneity. Notably, the image encoder converges less effectively to the unified BEV feature space than the LiDAR encoder, leading to underfitting. Consequently, the multimodal detector fails to fully leverage contextual information from the image modality, degrading overall performance. Moreover, this heterogeneity poses significant challenges in multimodal data preprocessing. While LiDAR-based detectors benefit from advanced 3D data augmentation techniques, such as ground-truth sampling and instance-level rotation, these methods struggle to maintain consistent transformations across LiDAR and image data in multimodal detection. The resulting imbalance in convergence and misalignment in augmentation further hinder the effectiveness of multimodal 3D detectors.

To address these challenges, we propose an Instance-level Contrastive Distillation (ICD) framework. In this framework, we use a pretrained LiDAR-based 3D detector as teacher to instruct the image encoder in our dual-branch network. Knowledge distillation is then applied to enhance the convergence of the image encoder towards the unified BEV space in joint optimization. Moreover, inspired by contrastive learning, we construct instance-level contrastive loss using ground-truth bounding boxes for more efficient distillation. Our approach facilitates fine-grained knowledge transfer between LiDAR and RGB images in the unified BEV space. This soft instance-level guidance encourages the 2D encoder to capture more robust 3D spatial transformations, reducing augmentation misalignment. Finally, to further enhance the performance of our model, we propose an inference-friendly multimodal fusion module based on linear attention. These novel methods empower our 3D object detector to achieve SOTA performance while running at 14 FPS on the multiclass 3D object detection task.

In summary, the contributions of this work are threefold:
\begin{itemize}
    \item We propose a novel Instance-level Contrastive Distillation framework for 3D object detection. Our framework achieves efficient knowledge distillation from single-modal to multimodal 3D detectors and incorporates object-aware contrastive learning to enable fine-grained modality alignment within the unified BEV space.
    \item We introduce a novel Cross Linear Attention Fusion Module (CLFM), which efficiently aggregates fair-sized multimodal BEV features, achieving computationally efficient inference with linear complexity.
    \item Our method achieves state-of-the-art (SOTA) performance on the KITTI multiclass 3D object detection benchmark while running at 14 FPS faster than the default LiDAR frame rate.
\end{itemize}

\section{Related Work}
\label{sec:related}
\subsection{BEV-based Multimodal 3D Object Detector}
Early works projected 3D point clouds onto 2D views~\cite{beltran2018birdnet} (e.g., depth images) to align BEV representations with images, but LiDAR-to-camera projections introduce geometric distortions. Transformer-based methods~\cite{li2024bevformer, zhou2024transfusion, zhao2021point} leverage flexible attention mechanisms to generate shared BEV features, while virtual-point approaches~\cite{xie2023cvpr, wu2023aaai, wu2023cvpr} enhance original point clouds via depth estimation networks. Though effective, these methods rely heavily on supervised depth estimation, limiting generalization. Alternative strategies fuse independently encoded image and LiDAR features in local RoIs~\cite{chen2017multi} or BEV space. BEVFusion~\cite{liu2023icra} adopts a dual-branch encoder with Lift-Splat-Shoot (LSS)~\cite{philion2020eccv}, predicting depth distributions to generate frustum point clouds. While this structure preserves image semantics and LiDAR spatial information, weak supervision in frustum generation leads to smearing effects in image-based semantic BEV features, degrading detection accuracy. To address this, our ICD framework enhances semantic-rich BEV consistency by leveraging robust LiDAR-based spatial features.

\subsection{Knowledge Distillation on 3D Object Detection}
Knowledge distillation has proven effective in transferring learned representations from a LiDAR teacher model to a camera student model, particularly in semantic-oriented tasks. For instance, BEVDistill~\cite{chen2022bevdistill} integrates dense feature distillation and sparse instance distillation to transfer knowledge from LiDAR features to a camera-based detector. UniDistill~\cite{zhou2023cvpr} aligns different modalities using predefined keypoints from ground truth bounding boxes and employs three distinct pointwise distillation strategies instead of a global approach. TiG-BEV~\cite{huang2022tig} introduces inter-channel and inter-keypoint feature distillation in the BEV space, reducing the cross-modal semantic gap and enhancing geometric learning for camera-based detection. Despite these advancements, the application of knowledge distillation for developing high-performance multimodal 3D object detectors remains an open research challenge.

\begin{figure*}[t]
  \centering
  \includegraphics[scale=0.44]{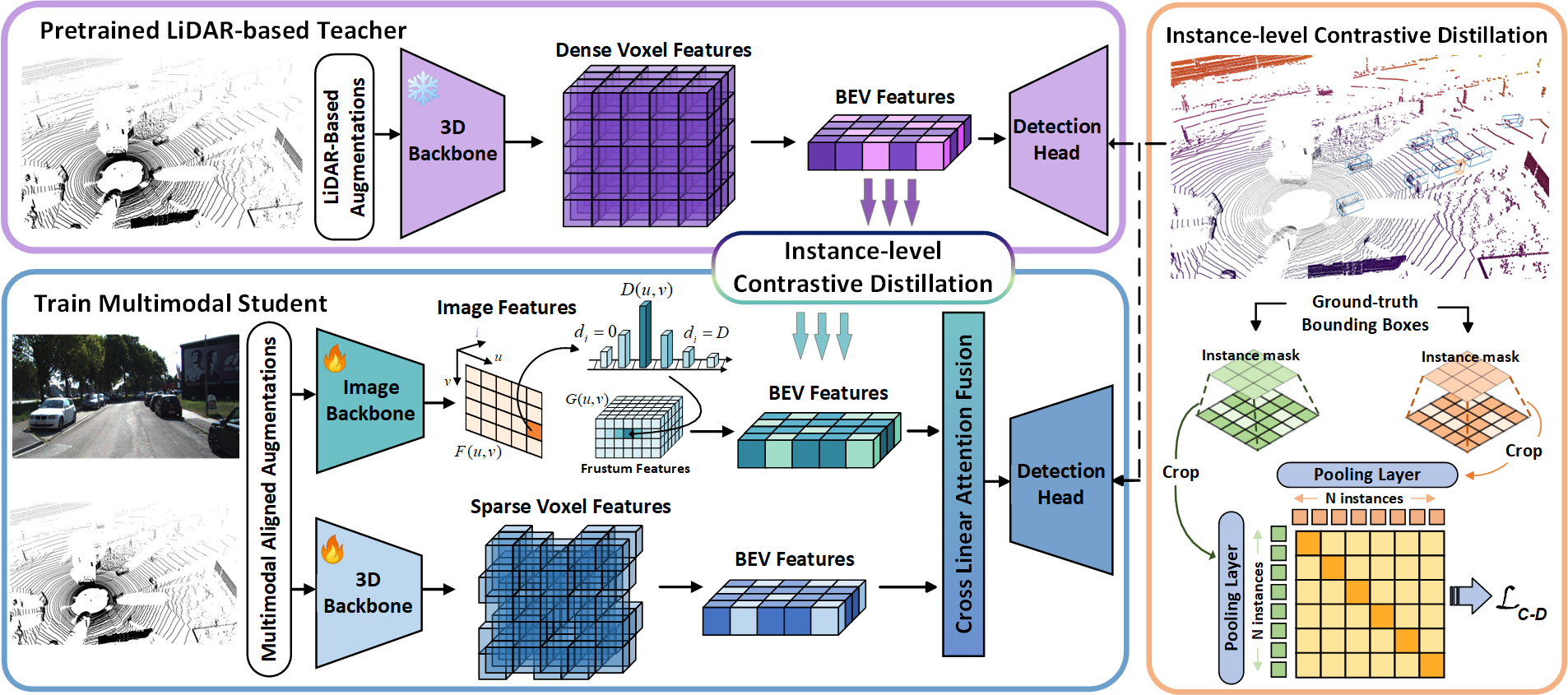}
  \vspace{-0.5cm}
  \caption{Overall architecture of our proposed Instance-level Contrastive Distillation (ICD) framework and Cross Linear Attention Fusion Module (CLFM). Firstly, a LiDAR-only teacher network is pretrained and its weights are frozen. Afterwards, we train the multimodal student with ICD and CLFM. The encoded features from the 3D branch and 2D branch are fully fused via our proposed CLFM, enabling high-performance online 3D object detection.}
  \vspace{-0.3cm}
  \label{fig:framework}
\end{figure*}

\subsection{Multimodal Fusion on 3D Object Detection}
Multi-sensor fusion is crucial in multimodal 3D object detection. Typical multimodal fusion methods predefine a unified space compatible with both modalities, such as BEV. Early BEV-based fusion methods concatenate features from different modalities within the BEV feature space, employing simple CNNs for scene-level fusion. Subsequent approaches, like TransFusion~\cite{zhou2024transfusion}, use image features to initialize object-aware queries in cross-attention, thus enabling proposal-level BEV feature fusion. More recently, IS-Fusion~\cite{yin2024cvpr} explores the synergistic interaction between scene and instance features and implements fusion strategies across multiple levels of granularity. However, these methods achieve high-performance multimodal feature fusion, their heavy reliance on attention mechanisms results in quadratic growth in computational complexity, significantly impacting the real-time performance of multimodal 3D object detectors.

Recently, the emergence of Mamba~\cite{gu2023mamba}, a state-space-inspired model with efficient inference capabilities, has garnered significant attention from researchers. Numerous studies have highlighted the similarities between Mamba’s core module, State Space Model (SSM), and attention mechanisms~\cite{vaswani2017nips}, leading to efforts to optimize the attention's computational complexity. For example, MLLA~\cite{han2025nips} has demonstrated remarkable results on benchmark tests by systematically comparing linear attention and SSM. Inspired by these insights, we propose a Cross Linear Attention Fusion Module that combines high performance with excellent inference speed. Our method can compute global dependencies across large-sized multimodal BEV features similar to cross-attention while maintaining competitive inference speed as ConvFusion. 
\section{Our Approach}
\label{sec:main}
The overall architecture of our proposed methods is shown in \cref{fig:framework}. We first employ aligned data augmentation strategies during preprocessing to ensure input consistency. At the training stage, we construct a dual-branch encoder to extract BEV features from images and LiDAR independently.  We utilize ICD for instance-level spatial knowledge transfer. The augmented point clouds are fed into the LiDAR-based teacher network to generate soft logits, guiding contrastive distillation in ICD. The fine-grained, aligned multimodal BEV feature map undergoes efficient global modeling via our CLFM module, producing well-fused representations. The fused features are then passed to the detection head for object detection, optimized through regression and classification losses, along with contrastive distillation loss from the ICD process. Each step of our method is detailed as follows.

\subsection{Preprocessing}
\label{sec:pre}
\begin{figure}[t]
  \centering
  \includegraphics[width=0.8\linewidth]{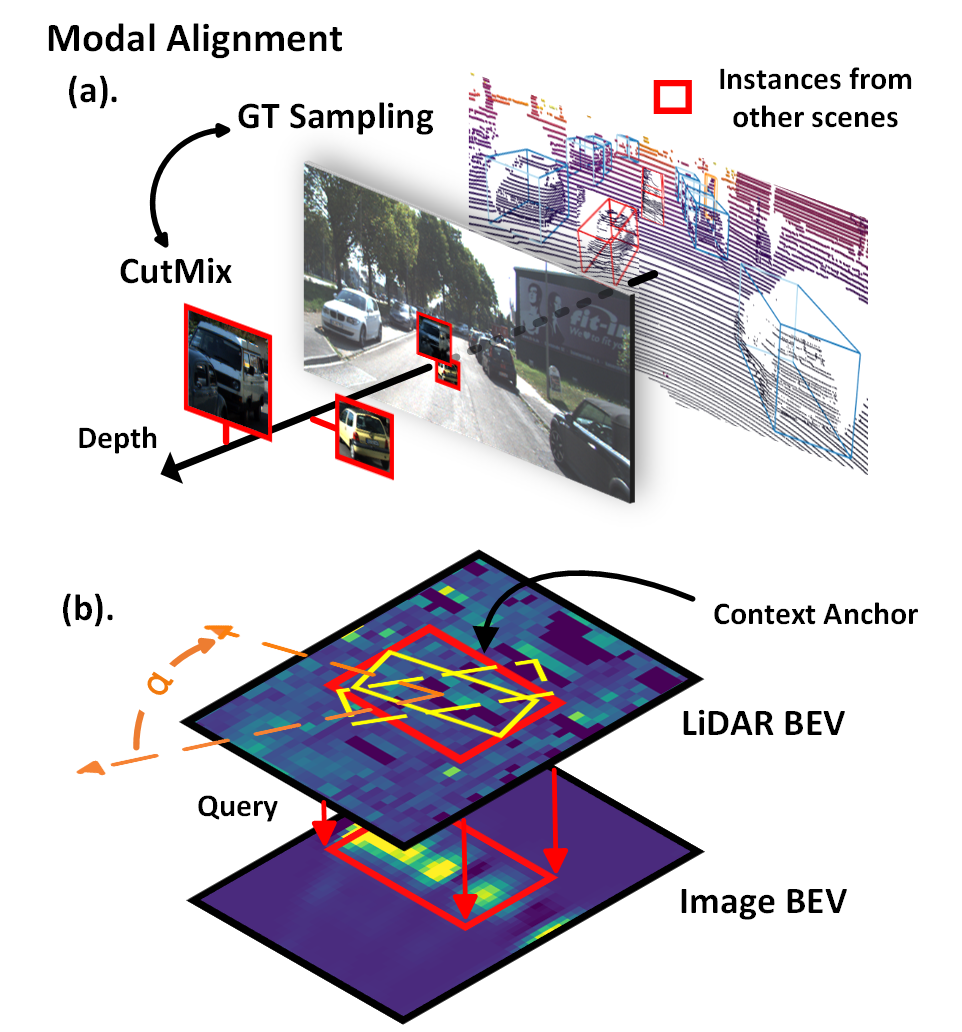}
  \vspace{-0.2cm}
  \caption{(a) CutMix and GT sampling ensure multimodal alignment at the input stage while preserving 2D perspective occlusion. (b) Rotated context anchors query image BEV features, enabling soft alignment of 3D rotations through instance-level feature distillation.}
  \label{fig:modelAlignment}
  \vspace{-0.3cm}
\end{figure}
\textbf{Augmentation alignment.} In data preprocessing, we apply LiDAR-based augmentations to the point cloud, including ground truth sampling, a widely used technique in LiDAR-based 3D detectors that enriches the density of target objects in the scene. To maintain preliminary alignment between images and point clouds, we adopt a depth-based CutMix~\cite{yun2019cvpr} approach, as illustrated in \cref{fig:modelAlignment}. This method leverages depth information from the point cloud space to determine the occlusion relationships between 2D instances during augmentation. In addition, the alignment of instance-level rotation augmentation in the point cloud is achieved through soft learning guided by the ICD process, as detailed in the following section.

\textbf{Multimodal 3D detector student.} In LSS-based multimodal 3D detectors, the RGB image and the voxelized LiDAR point clouds are input into the 2D and 3D backbone to obtain image feature $\mathrm{F}_\text{image} \in \mathbb{R^{H \times W \times C_{\text{l}}}}$ and LiDAR feature $\mathrm{F}_\text{lidar} \in \mathbb{R^{X \times Y \times Z \times C}}$, respectively. In the image branch, the extracted 2D feature $\mathrm{F}_\text{image}$ is first processed by a lightweight Depth Net $\phi(\cdot)$ to predict a depth distribution $p^{i}_\text{depth} \in \mathbb{R^D}$ for each pixel $i$, assisted by the projected depth map $\mathrm{F}_\text{depth}$. The semantic context of each feature pixel is then scattered into $D$ discrete points along the camera ray to generate frustum point clouds $\text{G}_\text{frustum} \in \mathbb{R^{H \times W \times D \times C}}$ with the predicted depth distribution. Finally, the frustum point clouds undergo BEV pooling to produce a semantically rich BEV feature map $\text{BEV}_\text{image}\in \mathbb{R^{H_{\text{b}} \times W_{\text{b}} \times C_{\text{i}}}}$ as:
\begin{align}
\mathrm{BEV}_\text{image} &= \mathrm{Pool}_\text{bev}(\mathrm{F}_\text{image} \times \phi(\mathrm{F}_\text{image},\mathrm{F}_\text{depth})).
\label{eq:frustum points}
\end{align}

In the LiDAR branch, for efficiency, we applied a dynamic voxel feature encoder to initially voxelize the point clouds, which only operates on non-empty voxels. Next, 3D sparse convolutions and 3D submanifold convolutions are applied to extract 3D features for preserving spatial sparsity. The encoded spatial 3D features are then aggregated by height compression to form the spatial BEV feature map. To align the BEV feature maps in the channel dimension ($C_{l} > C_{i}$) for the late fusion, $\mathrm{Conv}_{1\times1}$ is used to obtain $\mathrm{BEV}_\text{lidar} \in \mathbb{R^{H_{\text{b}} \times W_{\text{b}} \times C_\text{i} }}$:
\begin{equation}
\mathrm{BEV}_\text{lidar} = \mathrm{Conv}_{1\times1}(\mathrm{Pool}_\text{bev}(\mathrm{F}_\text{lidar})).
\label{eq:lidar BEV}
\end{equation}

\textbf{LiDAR-based 3D detector teacher.} For the LiDAR-based teacher network, we pretrain the Voxel-RCNN model~\cite{deng2021aaai}, which shares a similar architecture to the student 3D backbone. Unlike the LiDAR branch in the student model, voxelization and the backbone network are developed without lightweight or sparsity-preserving modifications to minimize the sparsity gap with $\mathrm{BEV}_\text{image}$. In training, point clouds with extensive augmentations $\mathrm{P}_\text{aug}$ are simultaneously fed into both the teacher model and the student 3D backbone. The parameters of the pretrained teacher network are frozen throughout the training process and its encoder outputs BEV features with the same channel dimensions as $\mathrm{BEV}_\text{image}$:
\begin{equation}
\mathrm{BEV}_\text{teacher} = \mathrm{Pool}_\text{bev}(\text{Encoder}_\text{teacher}(\mathrm{P}_\text{aug})).
\label{eq:teacher BEV}
\end{equation}

\subsection{Instance-level Contrastive Distillation}
In joint optimization, different convergence rates between modalities can cause one modality to dominate while the other remains underfitted. To tackle this, we introduce a knowledge distillation framework that enhances representation learning in the underfitted modality, ensuring balanced convergence and preventing multimodal detector performance degradation. 

However, objects in BEV representations are typically small and sparsely distributed. Directly computing a global distillation loss between dense $\mathrm{BEV}_\text{image}$ and sparse $\mathrm{BEV}_\text{teacher}$ results in unnecessary computations over large background areas, introducing noise. Moreover, the spatial context of instances eliminates false positive objects in the scene, enabling more precise modal alignment. Based on these considerations, we begin by generating 2D anchors from ground-truth bounding boxes ($bbox_\text{3D}$) to capture the local context of object instances:
\begin{equation}
\mathrm{Anchor}^i = \mathrm{max}_\text{x,y},\mathrm{min}_\text{x,y}\{\mathrm{Pool}_\text{bev}(bbox^i_{3D})\}.
\label{eq:anchor}
\end{equation}
Instance features are then cropped using the anchors from both $\mathrm{BEV}_\text{teacher}$ and $\mathrm{BEV}_\text{image}$. These features are pooled to a unified size, which enables efficient parallel processing:
\begin{align}
\mathbf{a}^{i} &= \mathrm{Pool}(\mathrm{Crop}(\mathrm{BEV}_\text{teacher},\mathrm{Anchor}^{i})),\\
\mathbf{b}^{i} &= \mathrm{Pool}(\mathrm{Crop}(\mathrm{BEV}_\text{image},\mathrm{Anchor}^{i})),
\label{eq:teacher BEV}
\end{align}
where, $\mathbf{a}^{i}$ and $\mathbf{b}^{i}$ denote the $i^{th}$ instance embeddings from $\mathrm{BEV}_\text{teacher}$ and $\mathrm{BEV}_\text{image}$ respectively. It is emphasized that the cropped instance features cannot be compressed through arbitrary pooling size, nor can they be regressed into a single feature value or vector. Such operations would severely degrade the spatial context of the BEV features. To investigate the influence of pooling size on detection performance, we conducted extensive ablation studies to identify the optimal pooling size in~\cref{sec:ablation}.

To achieve precise modality alignment and effectively supervise instance-level knowledge transfer, we incorporate the normalized temperature scaled cross-entropy loss~\cite{chen2020icml} into our knowledge distillation framework:
\begin{gather}
\text{sim}(\mathbf{a}^i, \mathbf{b}^i) = \frac{\mathbf{a}^i \cdot \mathbf{b}^i}{\|\mathbf{a}^i\| \|\mathbf{b}^i\|}, \\
\mathcal{L}_{\text{contrast}} = - \sum_{i \in \text{I}} \log \frac{\exp(\text{sim}(\mathbf{a}^i, \mathbf{b}^i) / \tau)}
{\sum_{j \in \text{I} \setminus \{i\}} \exp(\text{sim}(\mathbf{a}^i, \mathbf{b}^j) / \tau)},
\label{eq:NT-Xent}
\end{gather}
where $\text{sim}$(·) denotes the similarity matrix between instance embeddings from $\mathrm{BEV}_\text{teacher}$ and $\mathrm{BEV}_\text{image}$, $\text{I}$ represents all sample objects in a batch, $ \tau $ is the temperature scaling parameter, adaptively learned during distillation, $(\mathbf{a}^i,\mathbf{b}^i)$ and $(\mathbf{a}^i, \mathbf{b}^j)$ denote positive and negative sample pairs.

The distillation temperature is dynamically adjusted during training, where lower values promote fine-grained spatial learning. In ICD, querying image BEV features using $bbox_\text{3D}$ enhances the image encoder’s ability to capture 3D spatial rotations, facilitating soft alignment with rotation-augmented LiDAR instances.

\begin{figure}[t]
  \centering
  \includegraphics[width=0.95\linewidth]{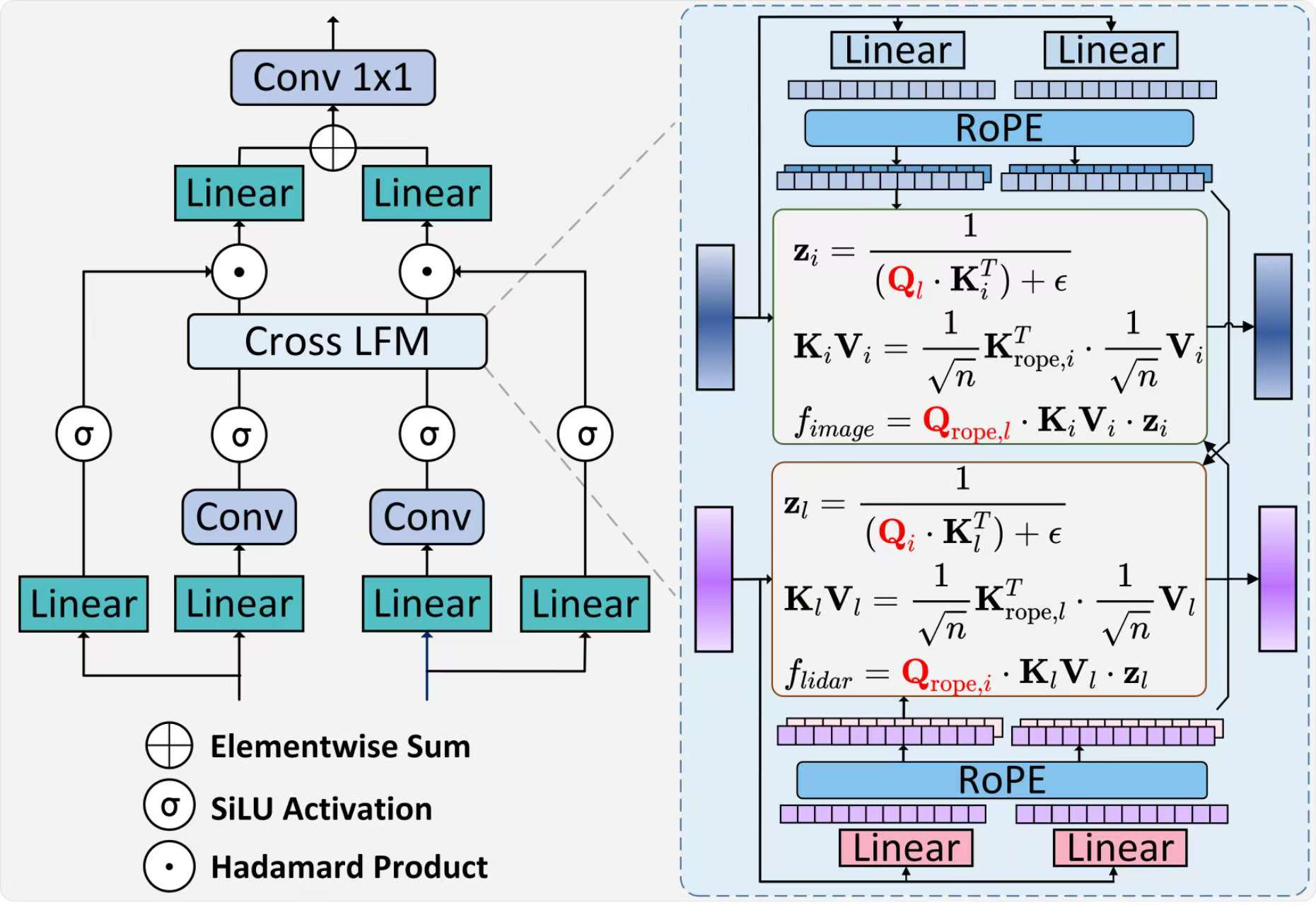}
  \caption{The architecture of Cross Linear Attention Fusion Module}
  \label{fig:parameval}
  \vspace{-0.3cm}
\end{figure}

\subsection{Cross Linear Attention Fusion}
In BEV-based 3D object detection, maintaining fair-sized BEV feature maps preserves spatial context but restricts the use of computationally intensive cross-attention for multimodal fusion. To overcome this limitation, we propose a linear attention-based fusion module that efficiently captures global dependencies with linear complexity.
As shown in~\cref{fig:parameval}, BEV features from both branches undergo linear projection $\text{Linear}(\cdot)$ and a convolution layer $\text{Conv}_{3\times3}(\cdot)$, followed by the $\text{SiLU}(\cdot)$ activation function for non-linearity before CLFM. For fusion, we apply linear projections $\text{Linear}(\cdot)$ to generate Queries, Keys, and Values for the flattened LiDAR BEV feature $\mathbf{x}$ and image BEV feature $\mathbf{y}$:
\begin{align}
\mathbf{Q}_l, \mathbf{K}_l, \mathbf{V}_l &= \text{Linear}(\text{SiLU}(\text{Conv}_{3\times3}( \text{Linear}(\mathbf{x})))),\\
\mathbf{Q}_i, \mathbf{K}_i, \mathbf{V}_i &= \text{Linear}(\text{SiLU}(\text{Conv}_{3\times3}( \text{Linear}(\mathbf{y})))).
\label{eq:1}
\end{align} 
Additionally, two extra linear projections perform channel transformations on $\mathbf{x}$ and $\mathbf{y}$, while a shortcut path preserving low-level features $\bar{\textbf{x}}$ and $\bar{\textbf{y}}$:
\begin{align}
\bar{\textbf{x}} = \text{Linear}(\text{SiLU}(\textbf{x})),
\bar{\textbf{y}} = \text{Linear}(\text{SiLU}(\textbf{y})).
\end{align}

Inspired by MLLA~\cite{han2025nips}, we employ $\text{elu}(\cdot)$ activation and $\text{RoPE}(\cdot)$ (Rotary Positional Encoding)~\cite{su2024neuroc} to process the queries and keys for both modalities $\mathbf{Q}_l, \mathbf{K}_l,\mathbf{Q}_i, \mathbf{K}_i$ to introduce the forget gate effects in long-range modeling. Then we approximate traditional self-attention by replacing $\mathrm{Softmax}(\cdot)$ operation with these low-rank kernel functions ($\text{elu}(\cdot)$, $\text{RoPE}(\cdot)$) in \eqref{eq:3}:  
\begin{gather}
\mathbf{Q}_{\text{rope}, \{l, i\}} = \text{RoPE}(\text{elu}(\mathbf{Q}_{\{l, i\}})), \\
\mathbf{K}_{\text{rope}, \{l, i\}} = \text{RoPE}(\text{elu}(\mathbf{K}_{\{l, i\}})), \\
\mathbf{K}_{\{l, i\}}\mathbf{V}_{\{l, i\}} = \frac{1}{\sqrt{n}} \mathbf{K}_{\text{rope}, \{l, i\}}^T \cdot \frac{1}{\sqrt{n}} \mathbf{V}_{\{l, i\}},
\label{eq:3}
\end{gather}
where $n$ denotes the number of heads. Our proposed fusion module employs a bidirectional fusion strategy, integrating image-to-LiDAR and LiDAR-to-image components to achieve effective and unbiased multimodal fusion. Taking image-to-LiDAR fusion as an example, shown as follows:
\begin{align}
\mathbf{z}_l &= \frac{1}{(\mathbf{Q}_i \cdot \mathbf{K}_l^T) + \epsilon},\\
\mathbf{K}_l\mathbf{V}_l &= \frac{1}{\sqrt{n}} \mathbf{K}_{\text{rope}, l}^T \cdot \frac{1}{\sqrt{n}} \mathbf{V}_l, \\
\hat{\textbf{x}}&= \mathbf{Q}_{\text{rope}, i} \cdot \mathbf{K}_l\mathbf{V}_l\cdot \mathbf{z}_l,
\label{eq:4}
\end{align}
where $\epsilon$ represents a small constant that prevents numerical instability by ensuring the denominator never reaches zero. $\mathbf{z}_l$ denotes the normalization factor for stabilizing the training. By leveraging the kernel functions, we can reduce computational cost from $O(n^2)$ to $O(n)$ to compute scalable long-range dependencies over multimodal BEV features. Finally, as shown in \eqref{eq:4}, we perform the image-to-LiDAR fusion by utilizing $\mathbf{Q}_{\text{rope}, i}$ to query $\mathbf{K}_l\mathbf{V}_l$. Similarly, we implement the LiDAR-to-image fusion in the same way. To integrate the low-level and fused features, we employ the Hadamard product $\odot$ and $\text{Linear}(\cdot)$. Finally, the LiDAR-to-image and image-to-LiDAR features are elementwise summed $\odot$, and processed through a $\text{Conv}_{1\times1}$ to facilitate cross-channel interaction, producing the fused BEV representation:
\begin{align}
\mathrm{F}_\text{fused}=\text{Conv}_{1\times1}(\text{Linear}(\hat{\textbf{x}} \odot \bar{\textbf{x}}) \oplus \text{Linear}(\hat{\textbf{y}} \odot \bar{\textbf{y}})).
\end{align}

\subsection{Loss functions}
The fused features are fed into detection head to predict final results. Our loss function is composed of $\mathcal{L}_{cls}$, $\mathcal{L}_{reg}$ and $\mathcal{L}_{contrast}$. The classification loss function $\mathcal{L}_{cls}$ is:   
\begin{equation}
    \scalemath{0.9}{
    \mathcal{L}_{\text{cls}} = -\alpha (1 - \hat{\beta})^\gamma \beta \log \hat{\beta} - (1 - \alpha) \hat{\beta}^\gamma (1 - \beta) \log (1 - \hat{\beta}),
    }
\label{eq:cls_loss}
\end{equation}
where \( \beta \) is the ground truth class label. \( \hat{\beta} \) is the predicted probability. \( \alpha \) is the weighting factor for positive samples. \( \gamma \) is the focusing parameter that reduces the weight of easy-to-classify samples. To refine the predicted 3D bounding box, we employ the Smooth L1 Loss to supervise its regression:
\begin{equation}
\scalemath{0.9}{
\mathcal{L}_{\text{reg}} = \sum_{i} \text{SmoothL1}(\Delta x_i, \Delta y_i, \Delta z_i, \Delta h_i, \Delta w_i, \Delta l_i, \Delta \theta_i),
}
\label{eq:reg_closs}
\end{equation}
\begin{equation}
\scalemath{0.9}{
\text{SmoothL1}(x) =
\begin{cases}
0.5 x^2, & |x| < 1, \\
|x| - 0.5, & \text{otherwise},
\end{cases}
}
\end{equation}
where \( (\Delta x, \Delta y, \Delta z) \) represent the center offsets, \( (h, w, l) \) are the box dimensions, and \( \theta \) denotes the yaw angle. Finally, integrated with the $\mathcal{L}_{contrast}$ from the ICD process, our overall loss function is formulated as follow:
\begin{equation}
\mathcal{L} = \mathcal{L}_{\text{cls}}+\mathcal{L}_{\text{reg}}+\mathcal{L}_{\text{contrast}}.
\label{eq:reg_closs}
\end{equation}

\section{Experimental Evaluation}
\label{sec:exp}
\subsection{Experimental Setup}
\subsubsection{Datasets and Metrics}
Following standard practices in 3D object detection, we conduct comprehensive experiments on the KITTI~\cite{Geiger2012cvpr} and nuScenes~\cite{caesar2020cvpr} benchmarks. 

\textbf{KITTI dataset.} 
The KITTI 3D object detection dataset consists of 7,481 LiDAR and image frames for training and 7,518 frames for testing. Following recent works, we split the training data into 3,712 frames for training and 3,769 frames for validation. For evaluation, we adopt the widely used 3D Average Precision (AP) metric under 40 recall thresholds (R40). The IoU thresholds in this metric are set to 0.7, 0.5, and 0.5 for cars, pedestrians, and cyclists, respectively.

\textbf{NuScenes dataset.} 
The nuScenes dataset is a large-scale autonomous-driving dataset for 3D detection and tracking, consisting of 700, 150, and 150 scenes for training, validation, and testing, respectively. Each frame contains one point cloud and six calibrated images covering the 360-degree horizontal FOV. For 3D detection, the main metrics are mean Average Precision (mAP) and nuScenes detection score (NDS). The mAP is defined by the BEV center distance instead of the 3D IoU, and the final mAP is computed by averaging over distance thresholds of 0.5m, 1m, 2m, 4m across ten classes. NDS is a consolidated metric of mAP and other attribute metrics, including translation, scale, orientation, velocity, and other box attributes.

\begin{table*}[h]
    \centering
    \caption{Performance Comparison on the KITTI Validation Set of 3D Object Detection Benchmark.}
    \label{tab:kitti_val_results}
    \renewcommand{\arraystretch}{1}
    \setlength{\tabcolsep}{7pt}
    \begin{tabular}{c|c|c|ccc|ccc|ccc|c}
        \toprule
        \multirow{2}{*}{\textbf{Method}} & \multirow{2}{*}{\textbf{Reference}} & \multirow{2}{*}{\textbf{Modality}} & \multicolumn{3}{c|}{\textbf{Car 3D (R40)}} & \multicolumn{3}{c|}{\textbf{Pedestrian 3D (R40)}} & \multicolumn{3}{c|}{\textbf{Cyclist 3D (R40)}} & \multirow{2}{*}{\textbf{mAP}}\\
        & & & Easy & Mod. & Hard & Easy & Mod. & Hard & Easy & Mod. & Hard \\
        \midrule
        PV-RCNN~\cite{shi2020cvpr}       & CVPR 2020 & L & 92.57 & 84.83 & 82.69 & 64.26 & 56.67 & 51.91 & 88.88 & 71.95 & 66.78 & 70.67 \\
        Voxel-RCNN~\cite{deng2021aaai}    & AAAI 2021 & L & 92.38 & 85.29 & 82.86 & 65.38 & 58.87 & 53.13 & 88.02 & 69.55 & 65.12 & 73.40 \\
        \midrule
        EPNet~\cite{huang2020eccv}         & ECCV 2020  & LC & 88.76 & 78.65 & 78.32 & 66.74 & 59.29 & 54.82 & 83.88 & 65.60 & 62.70 & 70.96 \\
        EPN++~\cite{liu2022pami}         & TPAMI 2022 & LC & 92.51 & 83.71 & 81.98 & 73.77 & 65.42 & 59.13 & 85.23 & 60.13 & 60.02 & 73.54 \\ 
        CAT-Det~\cite{zhang2022cvpr}       & CVPR 2022  & LC & 90.12 & 81.46 & 79.15 & \underline{74.08} & \underline{66.35} & 58.62 & 87.64 & 72.82 & 68.20 & 75.42 \\
        SFD~\cite{xie2023cvpr}           & CVPR 2022  & LC & 95.20 & \textbf{88.44} & \underline{85.99} & 72.78 & 64.46 & 58.19 & 88.85 & 72.27 & 67.86 & 77.12 \\
        LoGoNet~\cite{li2023cvpr}       & CVPR 2023  & LC & 92.04 & 85.04 & 84.31 & 70.20 & 63.72 & 59.46 & \underline{91.74} & \textbf{75.35} & \textbf{72.42} & 77.14 \\
        TED-M~\cite{wu2023aaai}         & AAAI 2023  & LC & \underline{95.55} & 86.48 & 84.26 & 72.69 & 65.02 & 58.29 & 91.64 & 72.02 & 67.50 & 77.05 \\
        VirConv-T~\cite{wu2023cvpr}     & CVPR 2023  & LC & \textbf{95.61} & \underline{87.98} & \textbf{86.64} & 73.06 & 66.15 & \underline{59.50} & 90.82 & 71.77 & 68.66 & \underline{77.79} \\
        Ours          & -          & LC & 92.57 & 85.72 & 83.37 & \textbf{76.60} & \textbf{71.97} & \textbf{66.89} & \textbf{93.50} & \underline{74.64} & \underline{70.22} & \textbf{79.59} \\
        \bottomrule
    \end{tabular}
\end{table*}

\begin{table*}[h]
    \centering
    \caption{Performance Comparison on the NuScenes Validation Dataset.}
    \label{tab:nuscenes_val_results}
    \renewcommand{\arraystretch}{1.2}
    \setlength{\tabcolsep}{6pt}
    \begin{tabular}{c|c|c c c c c c c c c c |c|c}
        \toprule
        \multirow{2}{*}{\textbf{Method}} & \multirow{2}{*}{\textbf{Modality}}    & \multicolumn{10}{c|}{\textbf{nuScenes Validation}} & \multirow{2}{*}{\textbf{NDS}} & \multirow{2}{*}{\textbf{mAP}}  \\
        & & Car  & Truck & C.V. & Bus  &Trailer& Barrier& Motor & Bike & Ped. & T.C. & & \\
        
        \midrule
        VoxelNeXt~\cite{chen2023cvpr}           & L    & 85.6 & 58.4 & 17.9 & 71.6 & 38.6 & 68.1 & 59.7 & 43.4 & 85.4 & 70.8& 67.1 & 60.3   \\
        TransFusion-L~\cite{zhou2024transfusion}       & L   & 87.2 & 59.8 & 26.0 & 73.3 & \underline{44.8} & 70.2 & 70.4 & 54.4 & 87.0 & 75.3 & 69.6 & 64.8  \\        
        \midrule
        TransFusion~\cite{zhou2024transfusion}         & LC  & 87.1 & \textbf{62.0} & 27.4 & \textbf{75.7} & 42.8 & \underline{73.9} & \textbf{75.4} & \underline{59.1} & \textbf{86.8} & \underline{77.0} & \underline{70.2} & \underline{66.7} \\
        BEVFusion~\cite{liu2023icra} (Baseline) & LC  & \underline{87.6} & 53.7 & \underline{28.5} & 73.8 & 41.9 & 72.4 & \underline{74.3} & \textbf{61.4} & \textbf{87.9} & \textbf{79.4} & 69.7 & 66.1  \\
        + ICD \& CLFM (Ours)                & LC  & \textbf{88.1} & \underline{61.3} & \textbf{34.2} & \underline{74.8} & \textbf{52.4} & \textbf{77.6} & 71.5 & 56.8 & 86.6 & 73.5 & \textbf{71.4} & \textbf{67.7}  \\  
        \bottomrule
    \end{tabular}%
\end{table*}

\subsubsection{Implemenation Details}
To demonstrate the univerisality and superiority of our ICD framework and CLFM, we conducted a single-view multimodal 3D detector based on Voxel-RCNN and a multi-view 3D detector based on BEVFusion~\cite{liu2023icra} for validations on KITTI and nuScenes, respectively. All comparative and ablation experiments were implemented with OpenPCDet~\footnote{https://github.com/open-mmlab/OpenPCDet} or MMDetection3D~\footnote{https://github.com/open-mmlab/mmdetection3d} libraries. The models were trained and evaluated on a workstation equipped with four NVIDIA RTX 4090 GPUs.

For the KITTI dataset, We adopted the same detection range, proposal number, and NMS threshold (0.3) as Voxel-RCNN~\cite{deng2021aaai} baseline detector. Following Voxel-RCNN, we set the voxel size to (0.05m, 0.05m, 0.1m). We trained the detectors with a batch size of two and an Adam optimizer with a learning rate of 0.01. For data augmentation, we used commonly used data augmentation strategies, including random flipping, global scaling with scaling factor $[0.95, 1.05]$ and global rotations about the Z axis between $[-\frac{1}{4} \pi,\frac{1}{4} \pi ]$. We pretrained vanilla Voxel-RCNN as the teacher model in our ICD framework and froze its weights during training the dual-branch multimodal student model. For the nuScenes dataset, we integrated our methods with BEVFusion using the same data augmentation strategies. Following VoxelNeXt~\cite{chen2023cvpr} , we set the voxel size to (0.075m, 0.075m, 0.2m). We first pretrained VoxelNeXt with a batch size of four and 20 epochs. Then we utilized it as the teacher model and trained a CLFM-modified BEVFusion with 30 epochs. 
\begin{figure*}[t]
  \centering
  \includegraphics[width=0.95\linewidth]{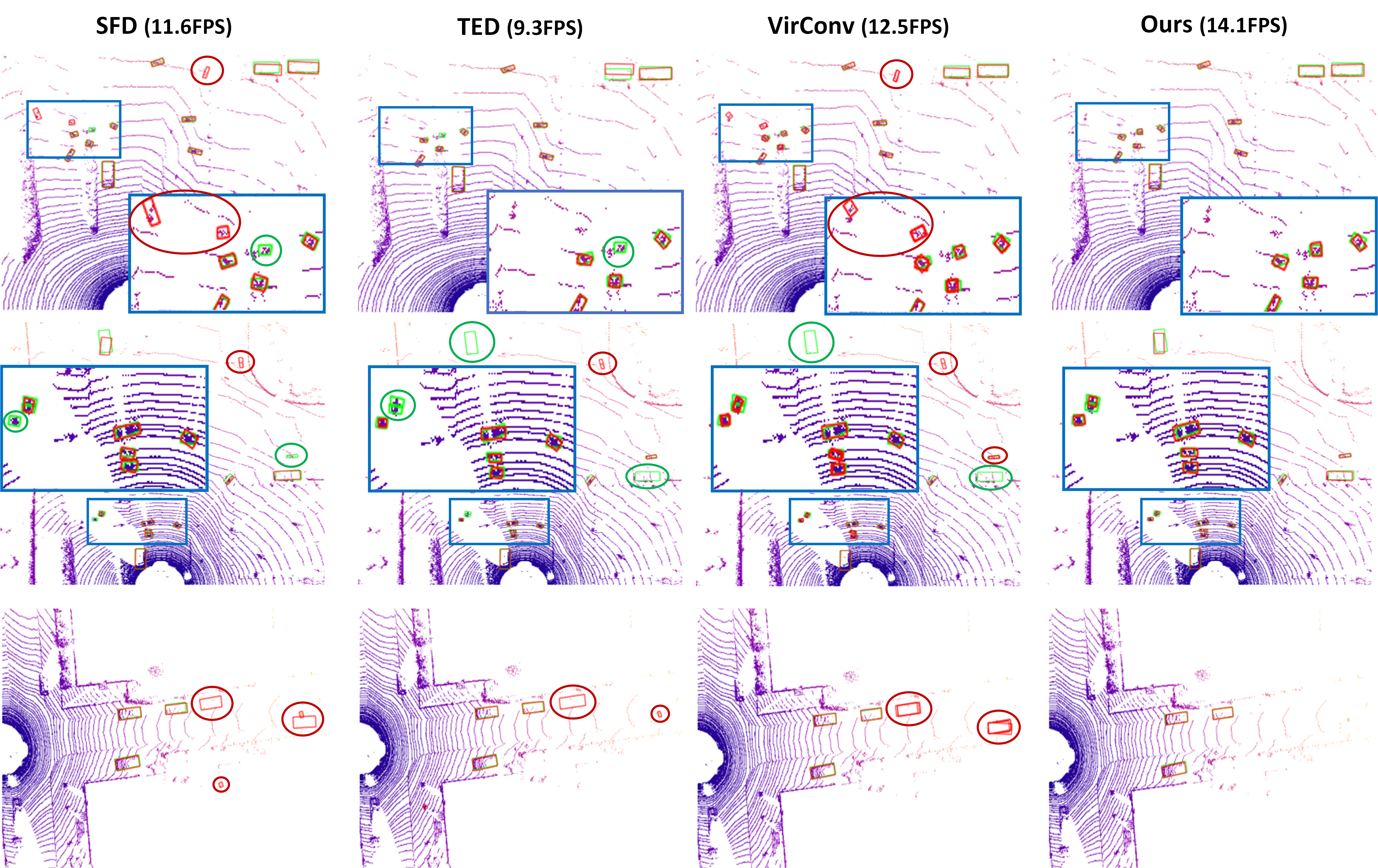}
  \caption{Qualitative results comparing our method with SOTA approaches on the multi-class KITTI 3D object detection task. \textcolor{green}{Green Box} represents the ground-truth bounding boxes; \textcolor{red}{Red Box} represents the predictions of methods; \textcolor{green}{Green Circle} represents the false negative objects; \textcolor{red}{Red Circle} represents the false postive predictions; \textcolor{blue}{Blue Anchor} is enlarged views to display tiny targets like pedestrians. Our method demonstrates strong detection performance on small targets while exhibiting robustness against false negatives and false positives.}
  \label{fig:bev}
\end{figure*}
\subsection{3D Object Detection Performance}
\textbf{Results on KITTI.} We conducted comparative experiments on the KITTI validation set, with results summarized in~\cref{tab:kitti_val_results}. In terms of the mean 3D average precision (mAP) for the multiclass 3D object detection task, our proposed method outperforms existing SOTA multimodal approaches by a substantial margin. In particular, it surpasses the Voxel-RCNN baseline by 6.19\% and outperforms VirConv-T~\cite{wu2023cvpr} by 1.60\%. Furthermore, our approach demonstrates remarkable performance in the Pedestrian and Cyclist categories. Specifically, when considering the moderate difficulty level, our method exceeds VirConv-T by 5.82\% and 2.87\% in the Pedestrian and Cyclist classes, respectively. Our proposed method exhibits excellent compatibility, as it neither relies on complex detection heads nor is restricted to specific 3D detector architectures. Additionally, it achieves online inference at a speed of 14 FPS while maintaining high performance with the simple Voxel-RCNN head.

\textbf{Results on nuScenes.} 
To demonstrate the universality of our proposed method, we also conducted comparative experiments on the nuScenes validation dataset, comparing our approaches with the baseline model BEVFusion. All reproduced experiments were implemented using the MMDetection3D library and executed on four NVIDIA RTX 4090 GPUs. As reported in \cref{tab:nuscenes_val_results}, our method enhances BEVFusion’s detection performance in terms of NDS and mAP, demonstrating that our methods are able to boost the multi-views 3D detector significantly. 

\subsection{Ablation Study}
\label{sec:ablation}
\textbf{Effectiveness of components.} We conducted a series of ablation experiments on the KITTI benchmark to evaluate the effectiveness of our proposed ICD framework and CLFM module. We implement a step-by-step ablation study, progressively refines a simple Voxel-RCNN model to derive our state-of-the-art solution. Initially, we replace the unimodal Voxel-RCNN encoder with a dual-branch (DB) encoder. Subsequently, a simple convolutional layer is employed to perform post-fusion of the semantic BEV features and spatial BEV features. As a result, it attains a performance of 75.5\% mAP, representing an improvement of 2.1\%. Subsequently, we investigated the impact of modality alignment on the detector's performance. Specifically, we examined two alignment strategies: (1) augmentation-based alignment (AA) during the data preprocessing stage, incorporating techniques such as CutMix and ground-truth sampling (corresponding to the third row of~\cref{tab:performance_comparison}), and (2) soft alignment through instance-level contrastive loss (corresponding to the fourth row of~\cref{tab:performance_comparison}).  Augmentation-based alignment improves mAP by 0.62\%, while soft alignment contributes a 2.77\% increase. In particular, combining both strategies resulted in a significant improvement of 3.57\%, highlighting the complementary benefits of these alignment techniques. Finally, we incorporate our CLFM module within the aligned BEV space to enable efficient multimodal fusion. This integration ultimately leads to SOTA performance, achieving a 3D mAP of 79.50\%.
\begin{table}[t]
    \renewcommand{\arraystretch}{1}
    \setlength{\tabcolsep}{16pt}
    \centering
    \caption{Ablation Studies of Different Modules.}
    \label{tab:performance_comparison}
    \begin{tabular}{ccccc}
        \toprule
        DB & AA & ICD & CLFM & 3D AP \\
        \midrule
         & & & & 73.40  \\
        \checkmark & & & & 75.50  \\
        \checkmark & \checkmark & & & 76.12  \\
        \checkmark & & \checkmark & & 78.27  \\
        \checkmark & \checkmark & \checkmark & & 79.07  \\
        \checkmark & \checkmark & \checkmark & \checkmark & \textbf{79.50}  \\
        \bottomrule
    \end{tabular}
\end{table}

\textbf{Different fusion schemes and pooling size.} We conducted a comprehensive study on commonly used post-fusion modules to evaluate the effectiveness of our proposed Cross Linear Attention Fusion Module (CLFM). The widely adopted fusion strategies can be categorized as follows: (1) employing self-attention (SA) or convolutional (Conv) layers to compute global or local correlations of multimodal data after concatenation, thereby achieving modal fusion;  (2) directly utilizing cross self-attention (CA) mechanisms to capture cross-modal relationships. Additionally, to examine the impact of different pooling sizes on the aggregation of local instance BEV features in contrastive distillation and its influence on the fusion strategy, we constructed three baseline models for each fusion strategy, employing pooling sizes of \(3\times3\), \(6\times6\), and \(9\times9\), respectively.
\begin{table}[t]
    \renewcommand{\arraystretch}{1}
    \setlength{\tabcolsep}{14pt}
    \centering
    \caption{Ablation Results by Using Different Fusion Schemes Under Three Pooling Size Baselines.}
    \label{tab:ablation_kitti}
    \begin{tabular}{c c c c}
        \toprule
        Fusion module & Pooling size  & 3D AP & FPS \\  
        \midrule
        \multirow{3}{*}{SA}   & \(3\times3\)   &      68.14     &\multirow{3}{*}{3.2}  \\
                              & \(6\times6\)   &      73.95    &                     \\  
                              & \(9\times9\)   & \textbf{75.67} &                     \\
        \midrule
        \multirow{3}{*}{CA}   & \(3\times3\)   &      75.65     &\multirow{3}{*}{5.7}  \\
                              & \(6\times6\)   &      76.79     &                     \\  
                              & \(9\times9\)   & \textbf{77.13} &                     \\
        \midrule
        \multirow{3}{*}{Conv} & \(3\times3\)   &      77.41     &\multirow{3}{*}{14.5}  \\
                              & \(6\times6\)   & \textbf{78.83} &                     \\  
                              & \(9\times9\)   &      78.27     &                     \\ 
        \midrule
        \multirow{3}{*}{CLFM} & \(3\times3\)   &      77.47     &\multirow{3}{*}{14.1}  \\
                              & \(6\times6\)   &      79.03     &                     \\  
                              & \(9\times9\)   & \textbf{79.59} &                     \\
        \bottomrule
    \end{tabular}
\end{table}
As shown in \cref{tab:ablation_kitti}, the experimental results demonstrate that our proposed CLFM consistently outperforms all three baseline tests, achieving superior performance across different pooling sizes. Moreover, CLFM attains an inference speed comparable to Conv layers. In contrast, the fusion strategy based on SA is susceptible to the local spatial aggregation method, resulting in performance fluctuations.
\begin{figure}[t]
  \centering
  \includegraphics[width=0.9\linewidth]{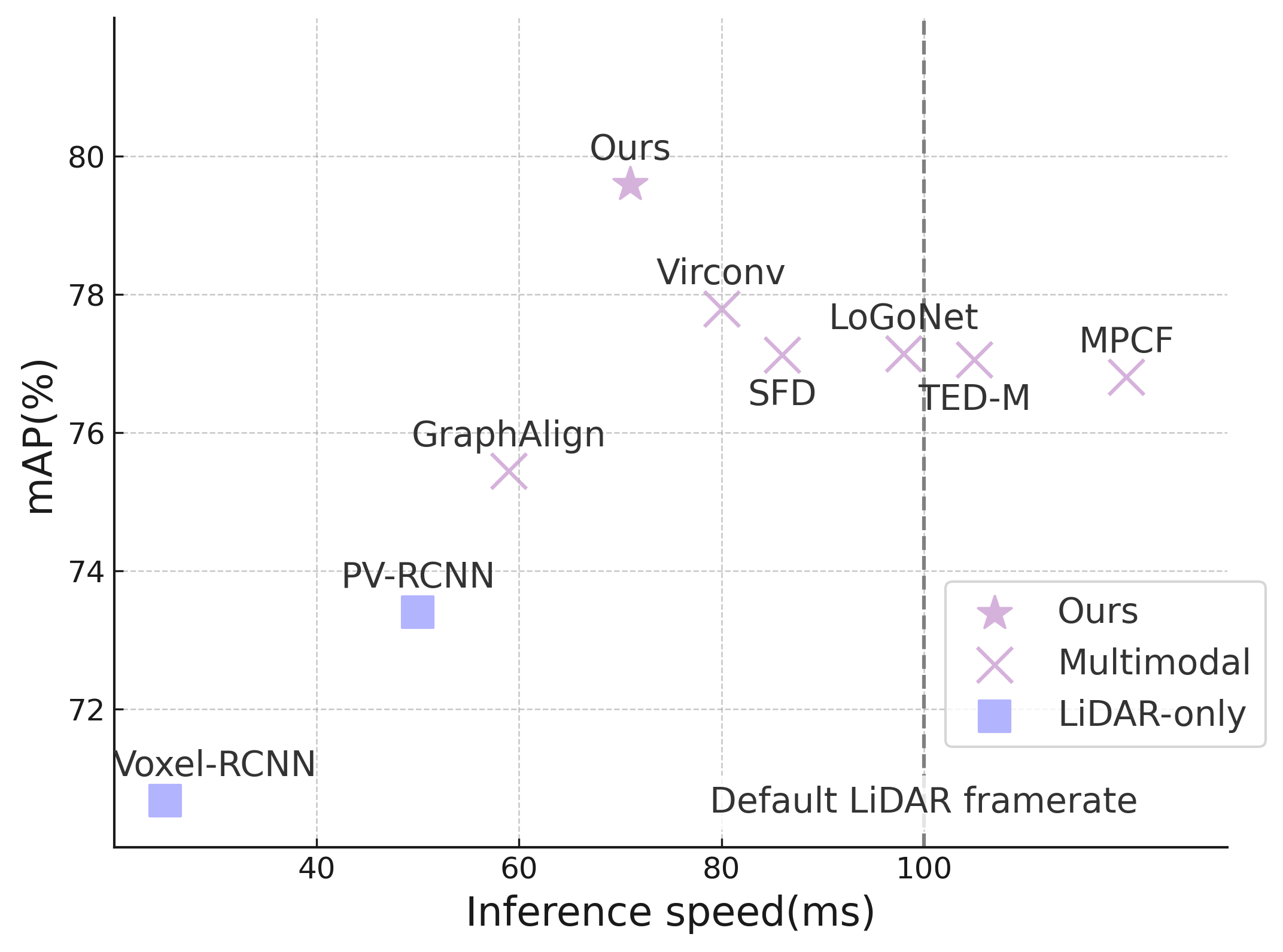}
    \vspace{-0.2cm}
  \caption{Our method achieves top mean average precision (mAP) on multi-classes 3D object detection in the KITTI benchmark and runs fast at 71ms. The dashed-line represents the default LiDAR framerate.}
  \label{fig:performance}
    \vspace{-0.3cm}
\end{figure}
\section{Conclusion}
\label{sec:conclusion}
In this paper, we propose a novel Instance-level Contrastive Distillation framework cooperated with a Cross Linear Attention Fusion module to construct an efficient and high-performance multimodal 3D object detector. The Instance-level Contrastive Distillation framework facilitates effective knowledge transfer between LiDAR and image modalities, ensuring fine-grained alignment in a unified BEV space. Meanwhile, the Cross Linear Attention Fusion module introduces a scalable and efficient fusion strategy, which enables the computation of long-range dependencies within BEV features at linear complexity.  To evaluate the effectiveness of our methods, we conducted extensive experiments on the KITTI and nuScenes datasets, along with a series of comprehensive ablation studies. Notably, our detector achieves SOTA performance at 14 FPS on the KITTI dataset without complicated detection heads. The experimental results highlight the strong synergy between modality alignment and multimodal fusion in 3D object detection task.


\bibliographystyle{ieeetr}

\bibliography{paper}

\end{document}